\documentclass[conference]{IEEEtran}
\IEEEoverridecommandlockouts
\usepackage{cite}
\usepackage{amsmath,amssymb,amsfonts}
\usepackage{bm}
\usepackage{algorithmic}
\usepackage{graphicx}
\graphicspath{{Images/}}
\usepackage{textcomp}
\usepackage{xcolor}
\def\BibTeX{{\rm B\kern-.05em{\sc i\kern-.025em b}\kern-.08em
    T\kern-.1667em\lower.7ex\hbox{E}\kern-.125emX}}
\usepackage[ruled, linesnumbered]{algorithm2e}
\usepackage[caption=false,font=footnotesize]{subfig}
\usepackage{tabularx,booktabs,multirow}
\newcolumntype{Y}{>{\centering\arraybackslash}X}
\newcolumntype{C}[1]{>{\centering\arraybackslash}m{#1}}

\usepackage{float}

\begin{document}

\title{Software-Level Accuracy Using Stochastic Computing With Charge-Trap-Flash Based Weight Matrix}

\author{
\IEEEauthorblockN{Varun Bhatt\IEEEauthorrefmark{1}, 
Shalini Shrivastava\IEEEauthorrefmark{2}, 
Tanmay Chavan\IEEEauthorrefmark{2} and
Udayan Ganguly\IEEEauthorrefmark{2}}
\IEEEauthorblockA{\IEEEauthorrefmark{1}
Department of Computing Science, University of Alberta, Edmonton, Canada \\
Email: vbhatt@ualberta.ca}
\IEEEauthorblockA{\IEEEauthorrefmark{2}
Department of Electrical Engineering, Indian Institute of Technology Bombay, Mumbai, India}
}

\maketitle

\begin{abstract}
The in-memory computing paradigm with emerging memory devices has been recently shown to be a promising way to accelerate deep learning. 
Resistive processing unit (RPU) has been proposed to enable the vector-vector outer product in a crossbar array using a stochastic train of identical pulses to enable one-shot weight update, promising intense speed-up in matrix multiplication operations, which form the bulk of training neural networks.
However, the performance of the system suffers if the device does not satisfy the condition of linear conductance change over around 1,000 conductance levels.
This is a challenge for nanoscale memories. 
Recently, Charge Trap Flash (CTF) memory was shown to have a large number of levels before saturation, but variable non-linearity.
In this paper, we explore the trade-off between the range of conductance change and linearity.
We show, through simulations, that at an optimum choice of the range, our system performs nearly as well as the models trained using exact floating point operations, with less than 1\% reduction in the performance. 
Our system reaches an accuracy of 97.9\% on MNIST dataset, 89.1\% and 70.5\% accuracy on CIFAR-10 and CIFAR-100 datasets (using pre-extracted features). 
We also show its use in reinforcement learning, where it is used for value function approximation in Q-Learning, and learns to complete an episode the mountain car control problem in around 146 steps.
Benchmarked to state-of-the-art, the CTF based RPU shows best in class performance to enable software equivalent performance.
\end{abstract}


\section{Introduction}
Deep Learning \cite{dl} has become the core driving force of artificial intelligence (AI). 
Applications such as image recognition, playing games, self-driving cars, and AI assistants are all made possible with the help of deep learning.
At the core of deep learning lies artificial neural networks (ANNs) \cite{perceptron}. 
ANNs are trained using large sets of data to approximate a function that explains the given data. 
Training is done using backpropagation \cite{backprop}, in which the weights of the neural network are updated based on gradient descent update rule.

The majority of the operations in training ANNs are matrix multiplications. 
Graphics processing units (GPUs) and Tensor processing units (TPUs) are specialized digital hardware designed to speed up this matrix multiplication. 
With faster computation cores, the bottleneck is currently in memory systems and data transfer \cite{gpu_mem}. 
Moreover, training ANNs for a typical real-world application requires hundreds of years of GPU time \cite{moravvcik2017deepstack}, leading to high energy costs.

In-memory computing \cite{le2018mixed} is an emerging paradigm, where data transfer is minimized by storing data and performing computation at the same place. 
Crossbar arrays with non-volatile memory have been shown to use lower energy, while also reaping the benefits of in-memory computation. 
Unfortunately, most of the devices struggle with precision and hence, the resulting performance of the system is not on par with their digital counterparts. 

Gokmen and Vlasov \cite{rpu} proposed a hypothetical resistive processing unit (RPU) that can be used to accelerate ANNs while being more energy-efficient than GPUs and having a negligible loss in accuracy. 
A crossbar architecture with a stochastic weight update rule allowed matrix multiplication in $\mathcal{O}(1)$ time. 
Linearity in weight update of the cross-point device and a high number of conductance levels were shown to be necessary to ensure good accuracy. 

Various approaches with nanoscale emerging memories like PCM \cite{nandakumar2018phase} and RRAM \cite{babu2018stochastic} have shown insufficient linearity to enable RPU as the sole memory. 
Recently, traditional charge trap flash memory has shown promising linearity \cite{agarwal2019using,shalini2019ultra}. 
However, their performance in the RPU framework has not been explored.

In this paper, we present a charge trap flash device that can act as a cross-point device in the RPU framework. 
We experimentally show a high number of conductance levels and approximately linear updates by choosing appropriate pulse width and voltage for weight update. 
Through simulations, we show that it indeed leads to a good accuracy when tested on MNIST, CIFAR-10 and CIFAR-100 datasets. 
In addition to supervised learning problems, we also successfully train a reinforcement learning agent on the Mountain Car environment.

\section{Related Work}

Matrix-vector multiplication and vector-vector outer product form the bulk of operations while training neural network.
RPU \cite{rpu} speeds up this computation using stochastic multiplication and hypothetical devices with linear weight updates.

Electronic synapses that have been proposed, such as nano-scale memristive synapses, may not have the gradual learning required for RPU.
Phase-change memory (PCM) based synapse has gradual positive conductance change, but abrupt negative conductance change, which requires novel synapse circuit design with enhanced controller complexity as well as a dual precision approach. 
Successful methods supplement weight storage in low precision but compact PCM with high precision but area inefficient CMOS based memory to achieve high performance \cite{suri2011phase,bichler2012visual,ambrogio2018equivalent,le2018mixed}.

With resistive random-access memories (RRAMs), multiple devices are required to obtain sufficiently gradual weight change to enable software equivalent learning \cite{boybat2018neuromorphic,shukla2018case}.
Additionally, RRAM (HfO\textsubscript{2}/PCMO/NbO\textsubscript{2}) and PCM based memory has additional process complexity / cost to be integrated into CMOS \cite{sze2017efficient}.

Floating-gate has been explored as an analog memory for neural networks extensively \cite{fujita1993floating}. 
However, horizontal floating-gate flash memory has been replaced by vertical charge trap flash memory with storage in silicon nitride traps for advanced technology nodes \cite{kang2016256}.


In contrast with memristor, a silicon-oxide-nitride-oxide (SONOS) based charge trap flash memory has significantly gradual conductance change with conductance saturation after 100 pulses \cite{agarwal2019using}. 
This may be compared to 20 pulses for PCM \cite{nandakumar2018phase}, or 20 pulses for PCMO based RRAM \cite{babu2018stochastic}. 
Maximum conductance change was between 5-20\% of the range of conductance and noise was around 5\%-10\% of the range of conductance. 
A dual precision approach in which one flash cell has a 1x factor and another has an 8x factor to define the weight was required to obtain software level accuracy on MNIST.
The weight updates also required varying pulse voltage and time, which would incur additional circuit costs.


Recently, a similar charge trap flash device has been programmed by quantum tunneling to show extremely gradual programming of 1,000-10,000 levels, which gives a 10-100x improvement over literature \cite{agarwal2019using}. 
The maximum conductance change per spike is controlled to $<$1\% of the range while the noise is 0.1\% of the range. 
However, linearity is not available in the entire range, which is essential for RPU applications. 
An important question is whether, by reducing the range of conductance, a smaller but more linear range can be found, which would enable software equivalent RPU, despite experimentally measured noise.

\section{Background}
\subsection{Artificial Neural Networks}

Artificial Neural Networks work based on the principle of multi-layered perceptron \cite[Chapter~6]{goodfellow2016deep}. 
Each layer of neurons performs a weighted linear combination of its inputs, applies a non-linear function, and passes the output to the next layer. 
Mathematically, given an input vector $\bm{x}$ and a weight matrix $\bm{W}$, a fully connected layer $i$ outputs 
\begin{equation}
    \label{eq:fp}
    \bm{a^{(i)}} = \phi^{(i)}(\bm{W^{(i)}} \bm{x^{(i)}}),
\end{equation}
where $\phi^{(i)}$ is some non-linear function called the activation. 
This operation is repeated for all layers, giving the output $\bm{\hat{y}}=f(\bm{x}, \bm{W})$.

In machine learning, neural networks are used to approximate the function between the input data and a target. 
Gradient descent is used to minimize a loss function ($\mathcal{L}(\bm{\hat{y}}, \bm{y})$) between the output of the neural network ($\bm{\hat{y}}$) and the true target ($\bm{y}$).
The gradients are calculated efficiently using backpropagation \cite{backprop}.

Backpropagation uses chain rule to propagate the gradients to the lower layers, given the gradients of the higher layers. 
Let $\bm{z^{(i)}} = \bm{W^{(i)}} \bm{x^{(i)}}$ and $\bm{\delta^{(i)}} = \frac{\partial \mathcal{L}}{\partial \bm{z^{(i)}}}$. 
Then,
\begin{equation}
    \label{eq:bp}
    \bm{\delta^{(i-1)}} = {\bm{W^{(i)}}}^T \bm{\delta^{(i)}} \odot {\phi^{(i-1)}}^\prime(\bm{z^{(i-1)}})
\end{equation}
\begin{equation}
    \label{eq:grad}
    \frac{\partial \mathcal{L}}{\partial \bm{W^{(i)}}} = \bm{\delta^{(i)}} {\bm{x^{(i)}}}^T
\end{equation}
where ${\phi^{(i-1)}}^\prime$ are the gradients of the activation functions and $\odot$ is the Hadamard (element-wise) product.
Equations \ref{eq:fp}, \ref{eq:bp}, and \ref{eq:grad}, along with the gradient descent update, form the core of training a neural network.

\subsection{Resistive Processing Unit}

Resistive processing units (RPUs) \cite{rpu} attempt to speed up the computation of the matrix-vector multiplication (Equations \ref{eq:fp}, \ref{eq:bp}) and vector-vector outer product (Equation \ref{eq:grad}).
For efficient hardware implementation, devices are arranged in a crossbar architecture with device conductance at each cross point representing a weight. 

First, Ohm's law, combined with Kirchhoff’s current law, is used to enable multiply-accumulate operation naturally in hardware. 
During forward pass (Equation \ref{eq:fp}), passing voltage proportional to $\bm{x^{(i)}}$ to the rows makes the current at the columns equal to the output of the layer $\bm{W^{(i)}} \bm{x^{(i)}}$.
Similarly, during backward pass (Equation \ref{eq:bp}), passing voltage proportional to $\bm{\delta^{(i)}}$ to the columns makes the current at the rows equal to ${\bm{W^{(i)}}}^T \bm{\delta^{(i)}}$, which is required for back-propagating the gradient.



\begin{figure}[t]
    \centering
    \includegraphics[width=0.8\columnwidth]{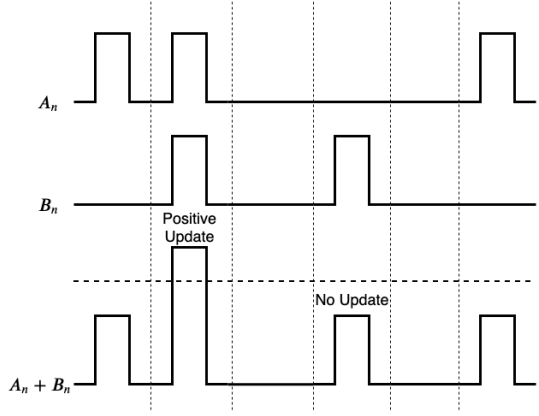}
    \caption{Analog multiplication using Stochastic pulse trains in RPU: Analog numbers are represented by a stochastic pulse train of identical pulses where the probability of high voltage in trains $A_n$ and $B_n$ is proportional to $x^{(i)}$, $\delta^{(i)}$ respectively. Updates occur at the coincidences, i.e., AND($A_n$, $B_n$), which have a probability proportional to $x^{(i)} \delta^{(i)}$. Polarity is reversed for negative updates.}
    \label{fig:rpu_pulse}
\end{figure}

Second, weight update by a simple stochastic AND operation is performed directly on non-volatile memory elements.
The outer product (Equation \ref{eq:grad}) is calculated using stochastic multiplication. 
Two pulse trains, with probability of high voltage proportional to $\bm{x^{(i)}}$, $\bm{\delta^{(i)}}$ respectively, are generated and passed through rows and columns respectively.
The voltage levels are set such that the resistive device updates its weight by $\Delta w$ when the pulses coincide, and there is no change when the pulses don't coincide. 
Since the expected number of coincidences is proportional to $x^{(i)} \delta^{(i)}$, the total weight update is proportional to the gradient in expectation.  
Figure \ref{fig:rpu_pulse} shows an example of pulse trains and the resulting update. 

The crossbar architecture and the stochastic weight update makes RPU more energy and area efficient compared to high precision digital multiplication blocks \cite{rpu}.

\section{Flash Synapse}
\subsection{Experimental Device}

We use a CTF capacitor (Figure \ref{fig:ctf}), which is fabricated as described by Sandhya et al. \cite{sandhya2008effect}. 
The device is fabricated on an n-Si substrate with 4 nm thermal SiO\textsubscript{2} as a tunnel oxide, 6 nm LPCVD Si\textsubscript{3}N\textsubscript{4} as charge trap layer (CTL), 12 nm MOCVD Al\textsubscript{2}O\textsubscript{3} as blocking oxide, and n+ polysilicon on 12” substrate by Applied Materials cluster tool.
Aluminum is used as a back contact. A self-aligned B implant and anneal is done to provide a source for minority carriers for fast programming as shown in Figure \ref{fig:ctf}a.


\subsection{Working as Synapse}

\begin{figure}[t]
    \centering
    \includegraphics[width=\columnwidth]{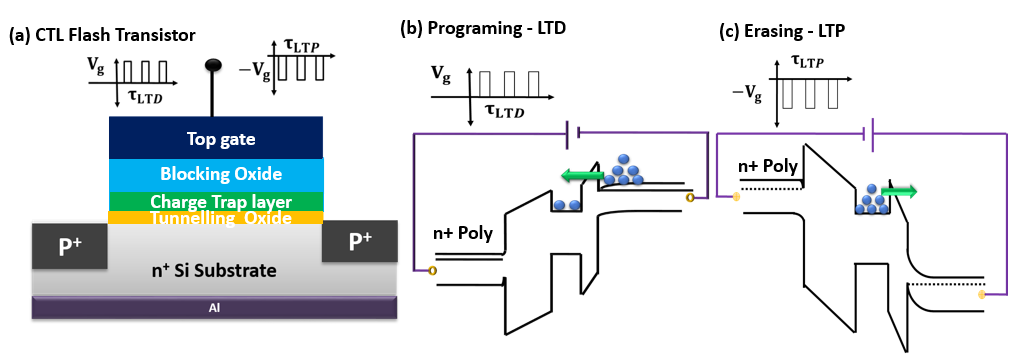}
    \caption{(a) Charge Trap Flash (CTF) schematic. Energy band diagram showing charge transport by quantum tunneling to charge/discharge silicon nitride atomic defects during: (b) Programming and (c) Erasing}
    \label{fig:ctf}
\end{figure}

The program/erase operation is based on FN tunneling. 
When a positive pulse is applied to the gate, electrons from the channel tunnel through the 4 nm tunnel oxide to be trapped in the CTL, i.e., programming (Figure \ref{fig:ctf}b). 
To erase, a negative pulse is applied to the gate. 
Electrons are ejected from the CTL by tunneling through the tunnel oxide (Figure \ref{fig:ctf}c). 

Programming and erasing results in a threshold voltage shift ($\Delta v_T$). 
The threshold voltage ($v_T$) is translated to drain current ($i_d$), which indicates the synaptic conductance ($g$) as follows:
\begin{align}
    i_d&=\beta_1(v_{GS}-v_T) v_{DS} \\
    i_d&=g \cdot v_{DS} \\
    \Delta g &\propto -\beta_2 \cdot \Delta v_T
\end{align}
where $\beta_1$ and $\beta_2$ are proportionality constants \cite{taur2013fundamentals}. 
Erasing ($\Delta v_T < 0$) results in potentiation ($\Delta g > 0$), while programming ($\Delta v_T > 0$) results in depression ($\Delta g < 0$). 
Henceforth, we use $v_T$ and $g$ interchangeably since they are simply the scaled version of each other.
An approximately linear and gradual change of conductance with the pulse number can be designed by pulse-width modulation \cite{shalini2019ultra}. 

\subsection{Experimental Data}
\label{sec:exp_data}

\subsubsection{Curve Fitting Device Updates}
\begin{figure}[t]
    \centering
    \subfloat[]{
    \includegraphics[width=0.47\columnwidth]{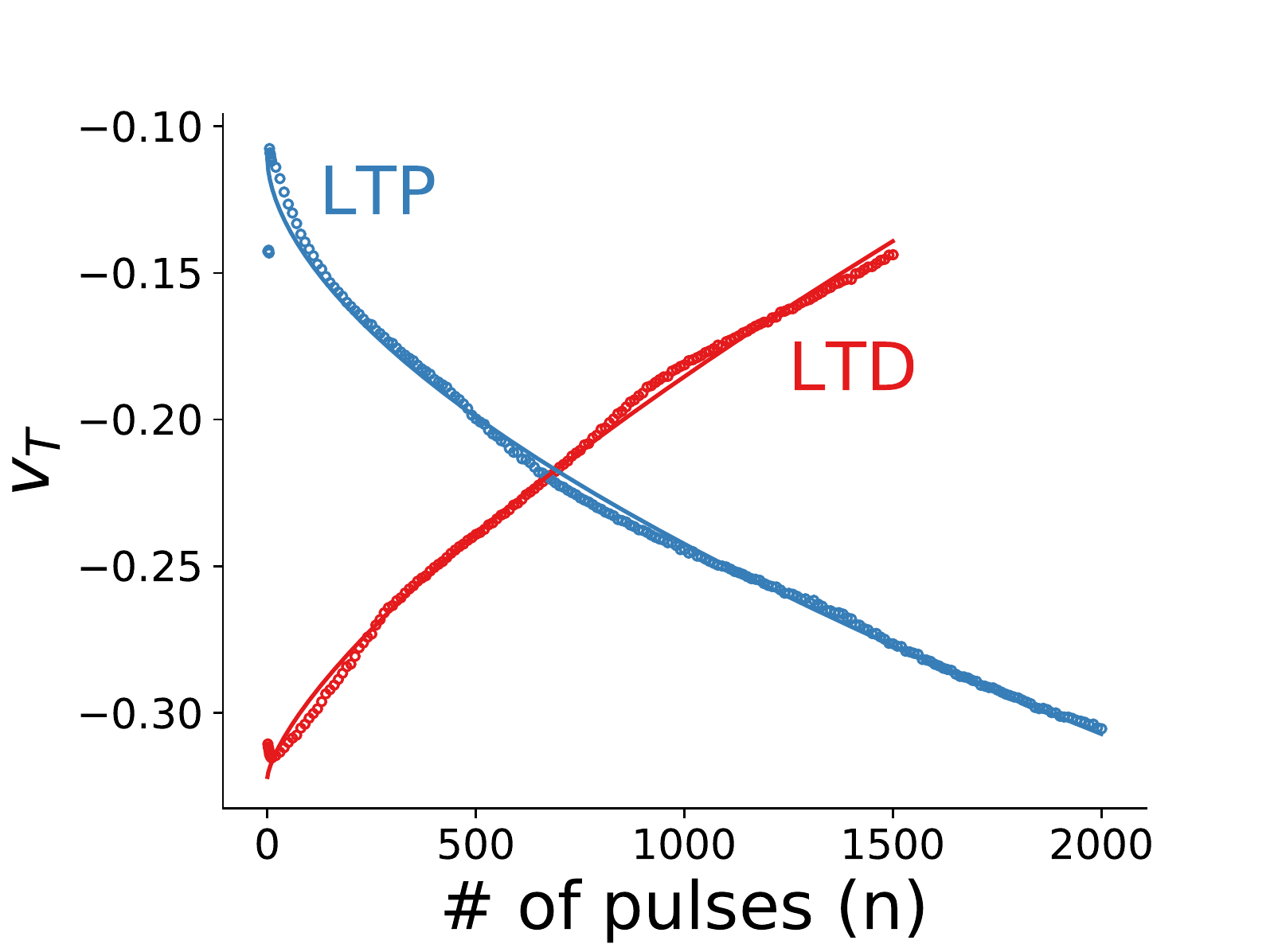}
    \label{fig:fit}
    }
    \hfill
    \subfloat[]{
    \includegraphics[width=0.47\columnwidth]{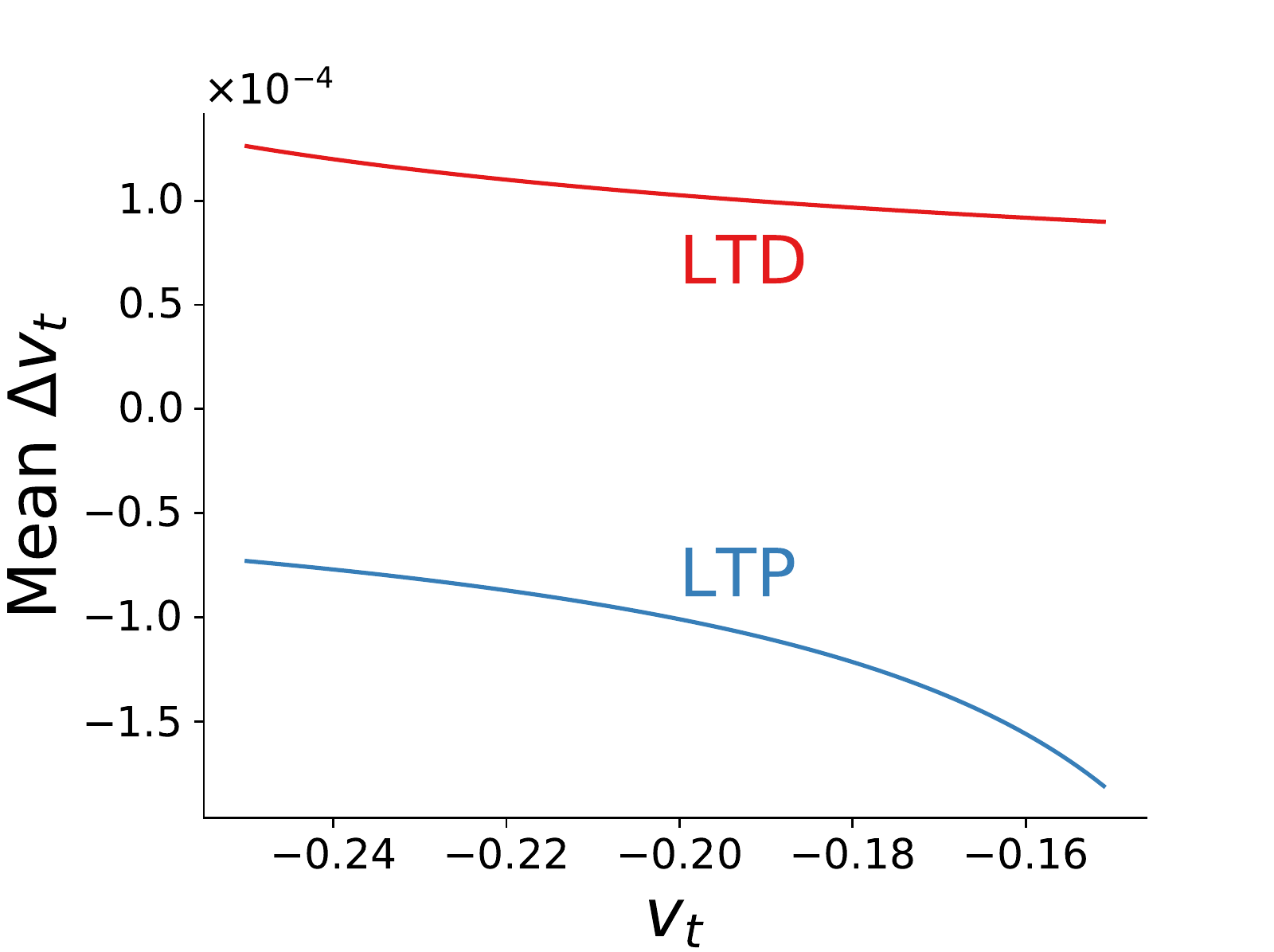}
    \label{fig:dvt}
    }
    \hfill
    \subfloat[]{
    \includegraphics[width=0.47\columnwidth]{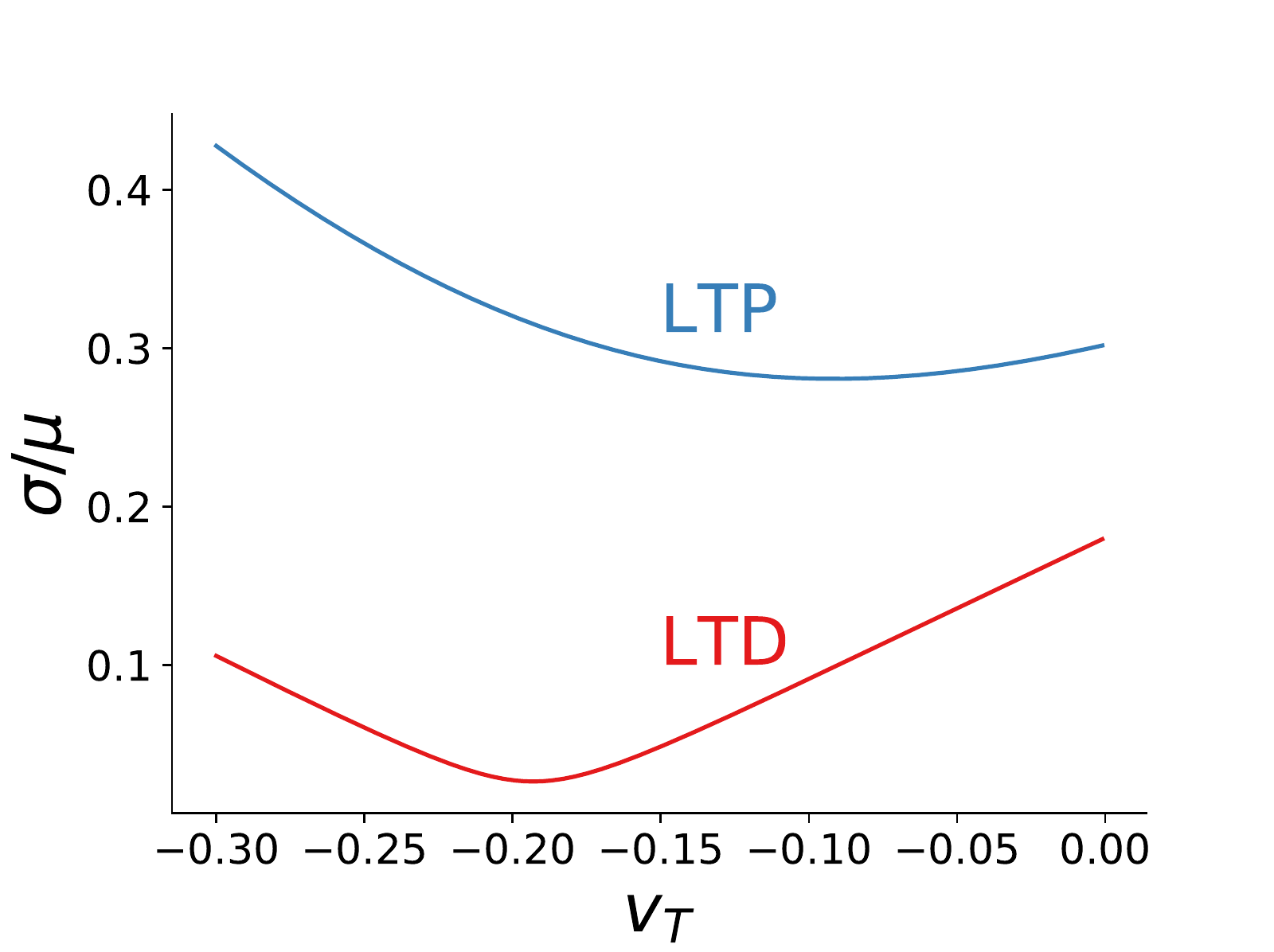}
    \label{fig:std}
    }
    \caption{(a) Experimental data (dots) and its curve fitting (lines) using the equation $v_T(n)=x_1(n)^{x_2} + x_3$. (b) Mean $\Delta v_T$ vs $v_T$ shows that $v_T$ shift is non-uniform.
    (c) Repeated measurements (6 times) of (a) is used to estimate the noise as a fraction of mean $\Delta v_T$ vs $v_T$. The experimental $\sigma / \mu$ is 30-40\% for LTP and $\sim$10\% for LTD.}
    \label{fig:my_label}
\end{figure}

We experimentally calculate the pulse amplitude and pulse width that gives an approximately linear weight change. 
Figure \ref{fig:fit} shows the experimental data of $v_T$ vs pulse number for LTD (using a pulse of +12.5V and 0.85ms width) and LTP (using a pulse of -12.5V and 15ms width). 
The scatter points are the observed data and the solid lines are the corresponding curve fits.

The curves were fit using the equation $v_T(n)=x_1(n)^{x_2} + x_3$ to minimize the mean squared error, with $x_1, x_2, x_3$ being the curve fit variables. 
The equation for $\Delta v_T$ was then found by setting $\Delta v_T (n) = v_T(n+1) - v_T(n)$ to get 
\begin{equation}
    \Delta v_T (v_T) = x_1 x_2 \left(\frac{v_T-x_3}{x_1}\right)^\frac{x_2-1}{x_2}
\end{equation}

We define $\Delta_0^+(g)$ as the positive change in $v_T$ when $v_T=g$ (using LTD data) and $\Delta_0^-(g)$ as the negative change in $v_T$ when $v_T=g$ (using LTP data).
Figure \ref{fig:dvt} shows the variation of $\Delta v_T$ with $v_T$.
The results of the curve fit gave $x_1, x_2, x_3 = 9.55 \times 10^{-4}, 7.19\times 10^{-1}, -3.22\times 10^{-1}$ respectively for LTD and $x_1, x_2, x_3 = -2.38\times 10^{-3}, 5.80\times 10^{-1}, -1.12\times 10^{-1}$ for LTP respectively, which implies that
\begin{align}
    \Delta_0^+(g) &= 4.50 (g + 0.32)^{-0.39} \times 10^{-5} \\
    \Delta_0^-(g) &= -1.74 (-g - 0.11)^{-0.72} \times 10^{-5}
\end{align}

\subsubsection{Characterization of Device Noise}
To find the noise in the updates, LTP and LTD experiments were repeated six times on the same device to characterize the variation within a device. 
For each experiment, a curve was fit and the corresponding $\Delta v_T$ was found. 
Then, for each $v_T$, the standard deviation ($\sigma$) of the evaluation of all six $\Delta v_T$ was found. 
Figure \ref{fig:std} shows the standard deviation as a percentage of mean vs $v_T$ for LTD and LTP.
This standard deviation is a measure of variation over time within a flash device - interpreted as noise.
To simplify the simulations, $\sigma$ was set to a high constant for all $v_T$ in our experiments.




\subsection{CTF in RPU array}
\label{sec:flash_rpu}

\begin{figure}[b]
    \centering
    \includegraphics[width=\columnwidth]{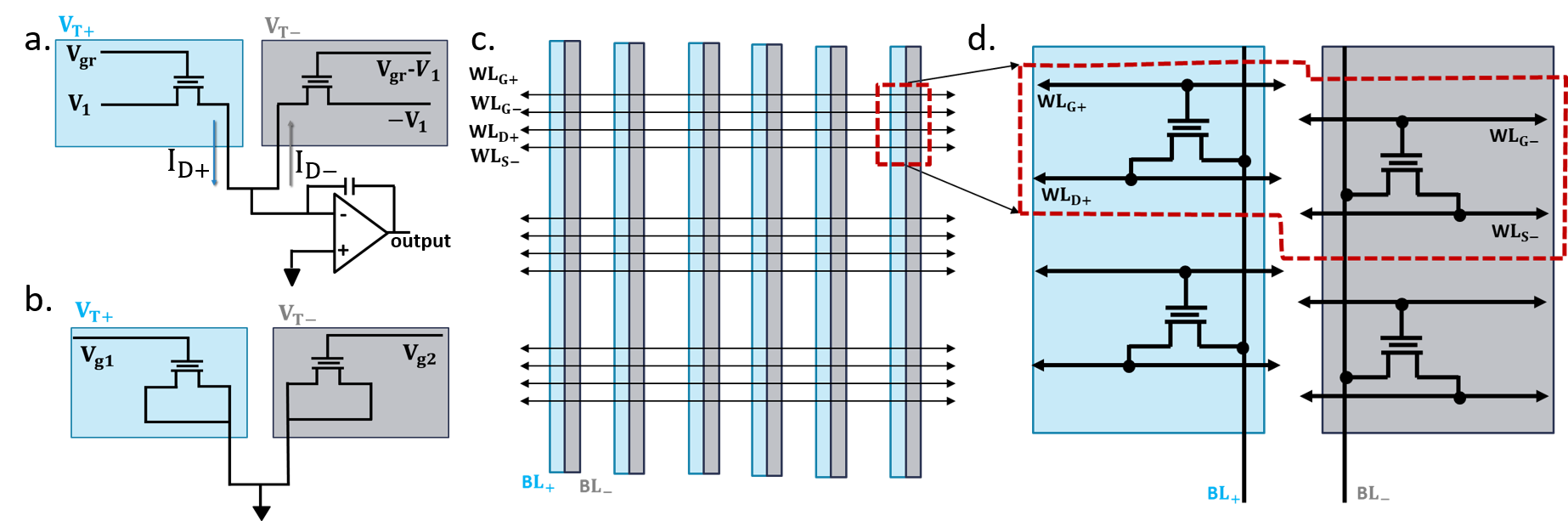}
    \caption{CTF in RPU array: Combination of two CTF used for representing $g_1$ (blue) and $g_2$ (grey). (a) The voltage applied at the gates generates currents at the source and drain respectively which are subtracted to produce weights capable of assuming positive and negative values. (b) Weight Update: Stochastic pulse trains $A_n$ and $B_n$ are applied to Gate and S/D shorted configuration respectively to produce an AND($A_n$,$B_n$) operation based voltage summation at the CTF device. (c) Crossbar architecture with bit line (BL) and word line (WL). (d) Each unit cell in the crossbar is the combination of two CTF.}
    \label{fig:circuit}
\end{figure}

\subsubsection{Simulating Device Updates}
The conductances of a CTF device are always positive, but the weights can be negative.
Thus, two devices are required to represent both positive and negative weights. 
Mathematically, the weight 
\begin{equation}
    w = k(g_1 - g_2)
    \label{eq:weight_scaling}
\end{equation} 
The scaling constant $k$ is used to control the range of device conductance.
In hardware, 2 CTF devices are arranged as shown in Figure \ref{fig:circuit}a. Applying voltages to the gates of the devices generates currents at the drain and source respectively. These currents are added to implement Equation \ref{eq:weight_scaling}.

$\Delta w$ is not constant since $\Delta g$ is a function of the current device conductances, and whether the update is positive or negative. 
The update is also noisy. 
Accommodating all these modifications, the positive and negative updates are given by 
\begin{align}
    \Delta^+ g &= \Delta_0^+(g) + N \\
    \Delta^- g &= \Delta_0^-(g) + N
\end{align}
where $N \sim \mathcal{N}(0, \sigma)$ is the noise.

\subsubsection{Controlling Linearity and Noise}

\begin{figure}[t]
    \centering
    \subfloat[]{
        \includegraphics[width=0.47\columnwidth]{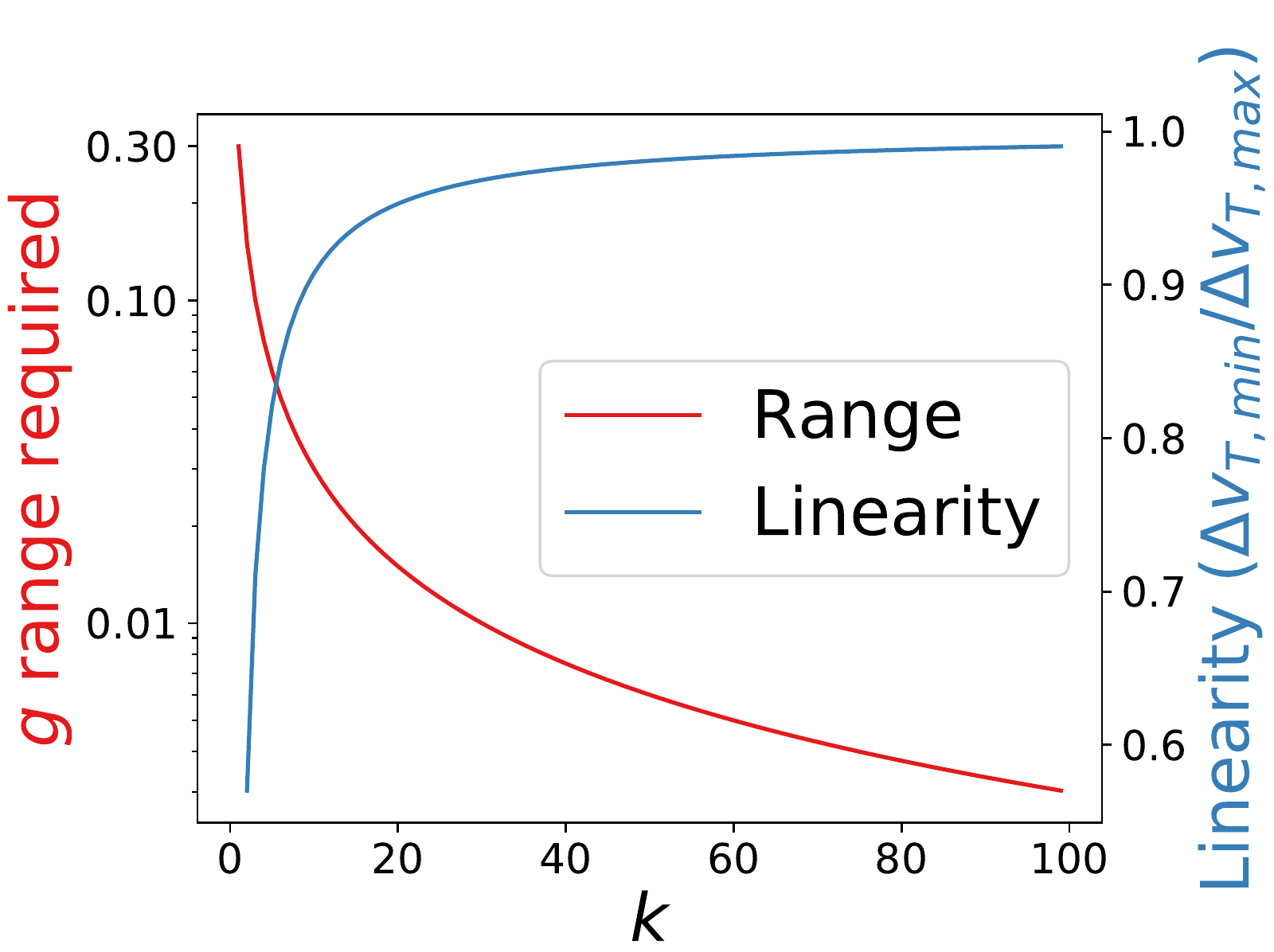}
        \label{fig:k_rl}
    }
    \hfill
    \subfloat[]{
        \includegraphics[width=0.47\columnwidth]{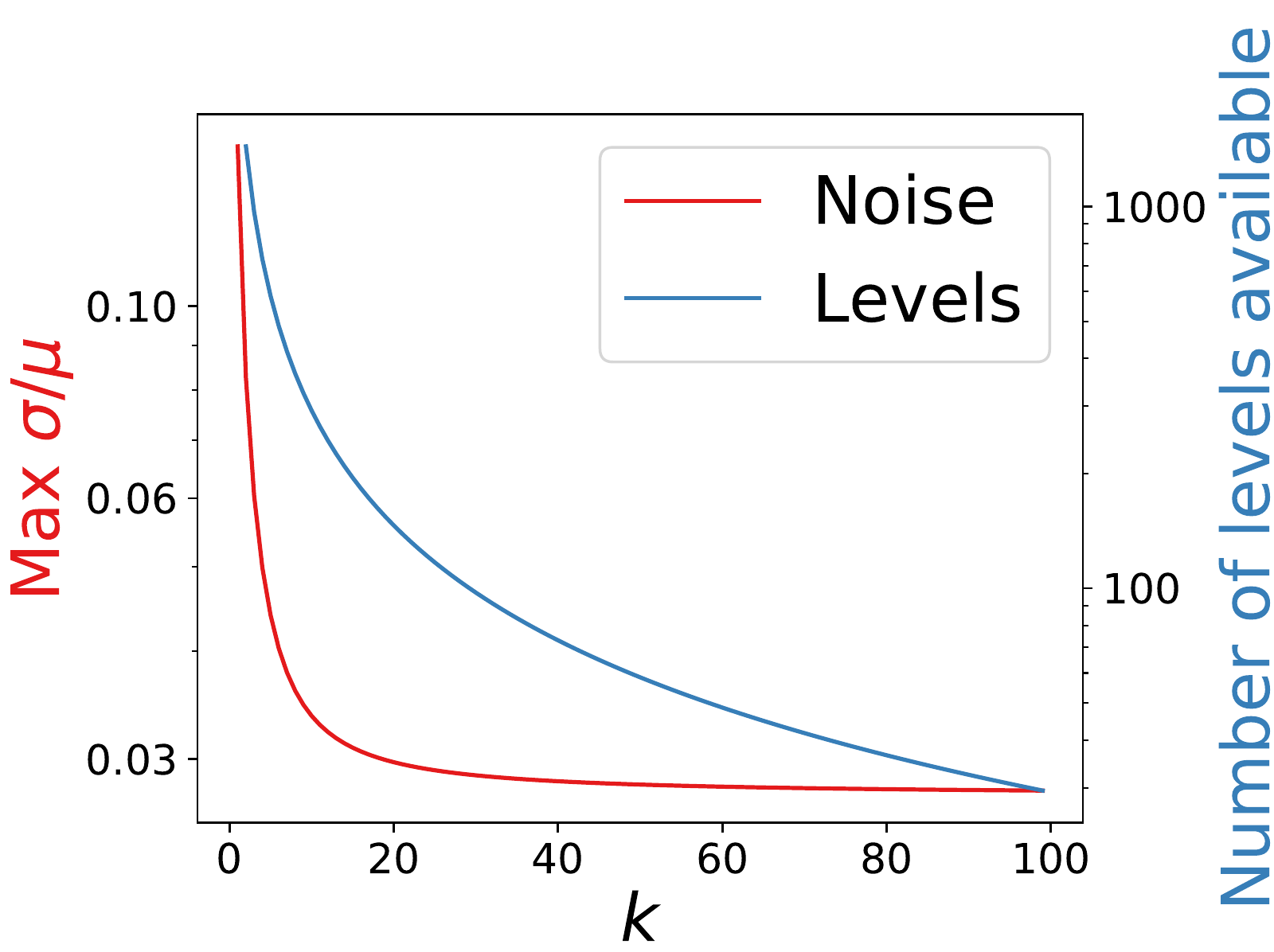}
        \label{fig:k_np}
    }
    \caption{Effect of $k$ on various device properties: (a) the required range of $g$ and linearity, (b) maximum standard deviation as a fraction of mean $\Delta v_T$ and the number of levels available. As $k$ increases, the required range of g is reduced. With appropriate centering, the range can be restricted to regions of high linearity and low noise. However, a smaller range also reduces the number of levels in the range.}
\end{figure}

Since the range of $w$ only depends on the dataset and the step size, $k$ controls the range of $v_T$ used, and hence, the noise, linearity, and the number of levels available. 
For example, Gokmen and Vlasov \cite{rpu} showed that the required range of $w$ was $(-0.3, 0.3)$, when training on MNIST dataset. 
Based on Equation \ref{eq:weight_scaling}, a conductance range of $\frac{0.3}{k}$ on each device is sufficient to represent this range.
Hence, a higher $k$ implies a lower required range of $g$, which can be observed in Figure \ref{fig:k_rl}.

Constraining $g$ to a lower range improves linearity (Figure \ref{fig:k_rl}). 
It also allows us to stay in the region with low noise, leading to lower maximum standard deviation as a fraction of mean $\Delta v_T$ (Figure \ref{fig:k_np}).
But as a trade-off, the number of levels available before it goes out of the range of $g$ is reduced (Figure \ref{fig:k_np}).
In Section \ref{sec:experiments}, we show the effect of this trade-off on the performance of the system.
In addition to the range, the center-point of the conductance range is optimized by trial and error to improve linearity.

\subsubsection{Circuit Design Considerations}
Performing an addition or subtraction of pulse trains is easier from a hardware perspective than an AND operation \cite{rpu}. 
To perform a positive update, two positive polarity pulse trains can be added such that a positive voltage pulse results at the coincidences. 
The polarities can be reversed to perform a negative update.
Since $\bm{x^{(i)}}$ and $\bm{\delta^{(i)}}$ are applied to the two ends of the crossbar, the polarity of the pulse trains must depend independently on the corresponding $x^{(i)}$ or $\delta^{(i)}$ and not the product $x^{(i)} \delta^{(i)}$. 
The input $x^{(i)}$ can be assumed to be positive since inputs are generally normalized between 0 and 1 and the common non-linear activations functions used in a neural network like sigmoid or ReLU only output positive values. 

\begin{figure}[t]
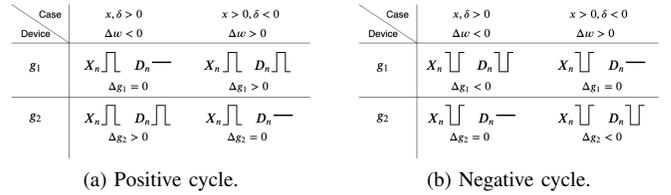

    \centering
    \subfloat[Positive cycle.]{
        \includegraphics[width=0.45\columnwidth]{Images/PositiveCycle.png}
        \label{fig:pos_update}
    }
    \hfill
    \subfloat[Negative cycle.]{
        \includegraphics[width=0.45\columnwidth]{Images/NegativeCycle.png}
        \label{fig:neg_update}
    }
    \caption{Two possible updates cycles where polarities of $X_n$ and $D_n$ can be set independently. Since $\Delta w \propto \Delta g_1 - \Delta g_2$, the updates resulting from these cycles are of the required polarity and magnitude. Any combination of these cycles (for example, alternating between them every iteration) is also valid.}
    \label{fig:update_cycle}
\end{figure}

Two possible update cycles with these constraints and the corresponding pulse polarities are shown in Figure \ref{fig:update_cycle}. 
We always use the positive cycle in our experiments.

Weight update in hardware for CTF devices is done by applying the voltage at the gate with respect to Source-Drain connected to the ground (Figure \ref{fig:circuit}b). 

As proposed by Gokmen and Vlasov \cite{rpu}, non-linear activation functions and their gradient can be implemented using an external circuitry. 
For the special case of ReLU activation, this external circuitry can be simplified. 
ReLU simply passes forward the positive inputs and blocks the negative inputs. 
The gradient is hence, 1 for positive inputs and 0 for negative inputs.

Figure \ref{fig:circuit}c shows the crossbar architecture with 4 word lines (WL) for applying voltages and 2 bit lines (BL) to read the currents. Figure \ref{fig:circuit}d shows a single unit cell in the crossbar with 2 CTF devices. Algorithm \ref{alg:flash_grad} describes the steps for calculating the weight update while simulating a CTF device.







\begin{algorithm}
\DontPrintSemicolon
\KwIn{Gradients ($\bm{\delta^{(i)}}$); Inputs ($\bm{x^{(i)}}$); Length of pulse trains ($PL$); Input scaling constant ($C$); Weight update functions $\Delta_0^+$, $\Delta_0^-$; Device conductances $\bm{G_1^{(i)}}$ and $\bm{G_2^{(i)}}$; Noise ($\sigma$).}
\KwOut{Updated values of the device conductances of the layer $\bm{G_1^{(i)}}$, $\bm{G_2^{(i)}}$.}

\For{each cross-point}{
    Let $g_1$, $g_2$ be the device conductances \;
    Find $x^{(i)}$, $\delta^{(i)}$ corresponding to the cross point \;
    Sample $X_1, \cdots, X_{PL} \sim Bernoulli(|C x^{(i)}|)$ \;
    Sample $D_1, \cdots, D_{PL} \sim Bernoulli(|C \delta^{(i)}|)$ \;
    Set the polarity of all $D_n$ equal to the sign of $\delta^{(i)}$ \;
    \For{each coincidence in $X_n \land D_n$}{
        \label{alg:for_coincidence}
        Sample noise $N \sim \mathcal{N}(0, \sigma)$ \;
        \uIf{$\delta^{(i)} < 0$}{
            $\Delta^+ g_1 \leftarrow \Delta_0^+(g_1) + N$ \;
            $g_1 \leftarrow g_1 + \Delta^+ g_1$
        }\Else{
            $\Delta^+ g_2 \leftarrow \Delta_0^+(g_2) + N$ \;
            $g_2 \leftarrow g_2 + \Delta^+ g_2$
        }
    }
}

\caption{Update calculation in CTF device simulation.}
\label{alg:flash_grad}
\end{algorithm}

\section{Experiments and Results}
\label{sec:experiments}
To test the performance of a neural network with flash synapse as the cross point device, we performed three experiments. 
We trained neural networks for supervised classification of digits in the MNIST dataset \cite{mnist}, images in the CIFAR dataset \cite{cifar}, and for reinforcement learning in the Mountain Car environment \cite{mountain_car}. 
All neural network operations were performed by simulating CTF devices as described in section \ref{sec:flash_rpu}. 
As a baseline in all the experiments, we performed the neural network training using exact floating point operations.  



Table \ref{tab:param} shows the list of hyperparameters used in the experiments. 
A combination of manual tuning and grid search was used to find these hyperparameters. 
Hyperparameters related to the CTF device and RPU were kept constant for all the experiments.

\begin{table}[t]
    \centering
    \caption{Hyperparameters.}
    \label{tab:param}
    \begin{tabularx}{\columnwidth}{cY}
        \toprule
        \textbf{Hyperparameter} & \textbf{Value} \\
        \midrule
        \multirow{3}{*}{Update step size ($\alpha$)} & MNIST: 0.01 \\
        & CIFAR: 0.1 \\
        & Mountain Car: 0.00625 \\
        \midrule
        Initial weights ($w_0$) & Kaiming uniform \cite{kaiming} \\
        \midrule
        Weight scaling factor ($k$) & $600 \alpha$ \\
        \midrule
        Initial device conductance ($g_{1,0}$, $g_{2,0}$) & $-0.2 \pm \frac{w_0}{2k}$ \\
        \midrule
        Pulse train length $PL$ & 10 \\
        \midrule
        Input scaling factor $C$ & $\sqrt{\frac{\alpha}{PL \cdot \Delta^+ g(c) \cdot k}}$ \\
        \bottomrule
    \end{tabularx}
\end{table}


\subsection{MNIST}

MNIST dataset consists of 60,000 training and 10,000 test images of 10 handwritten digits, each of size 28x28 pixels. 

A fully connected neural network with 2 hidden layers consisting of 256 and 128 neurons respectively, was used for classification. 
The neural network was trained for 10 epochs. 
Experiments were repeated 10 times with different random seeds and the train accuracy was recorded after every 5,000 images. 
The test accuracy was also recorded after every 5,000 training images by performing classification on the complete test set. 
Two sets of experiments were performed, with noise standard deviation ($\sigma$) being 10\% of the mean in one and 100\% of the mean in the other. 

\begin{figure}[t]
    \centering
    \subfloat[]{
    \includegraphics[width=0.47\columnwidth]{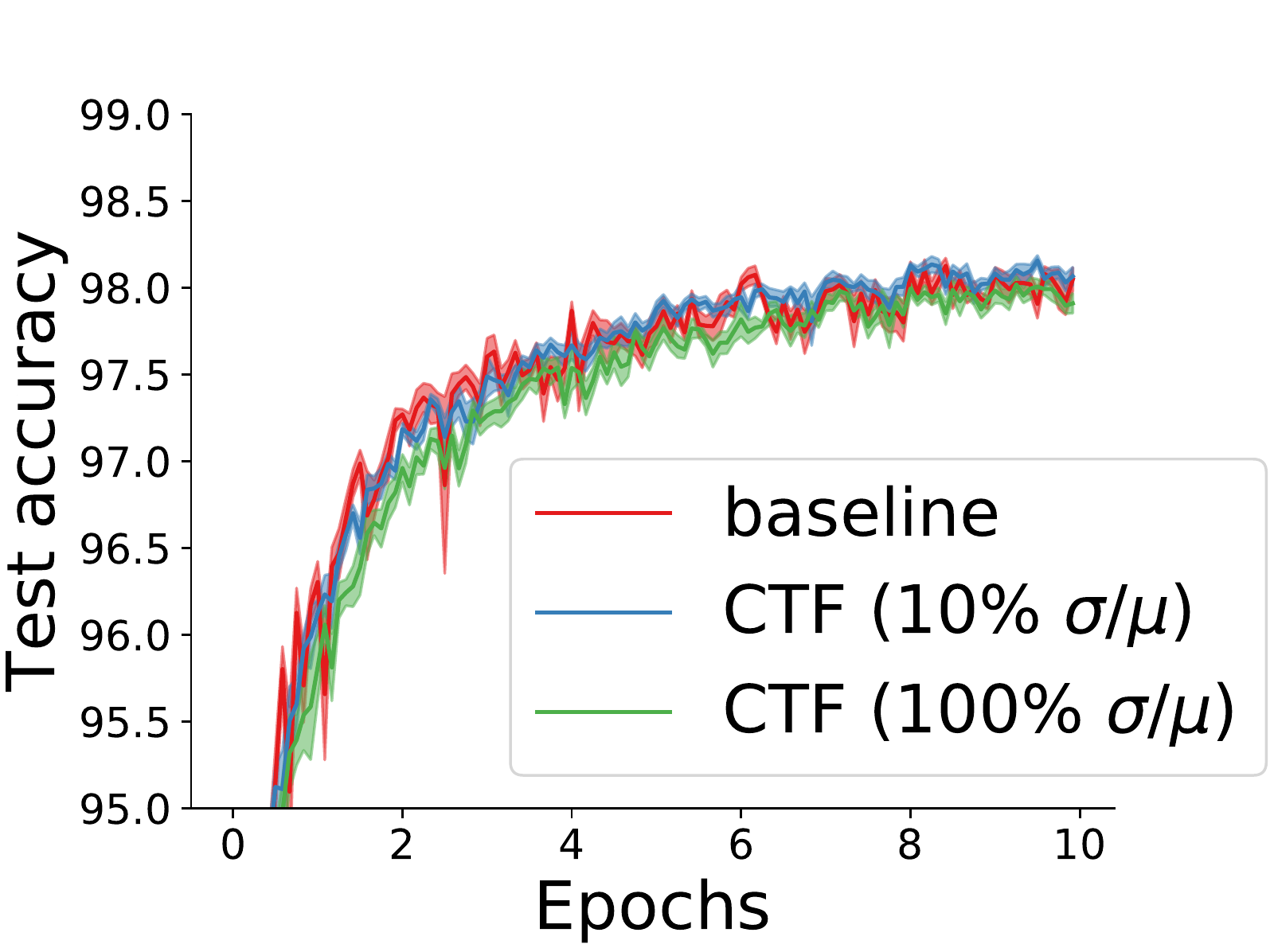}
    \label{fig:mnist}
    }
    \hfill
    \subfloat[]{
    \includegraphics[width=0.47\columnwidth]{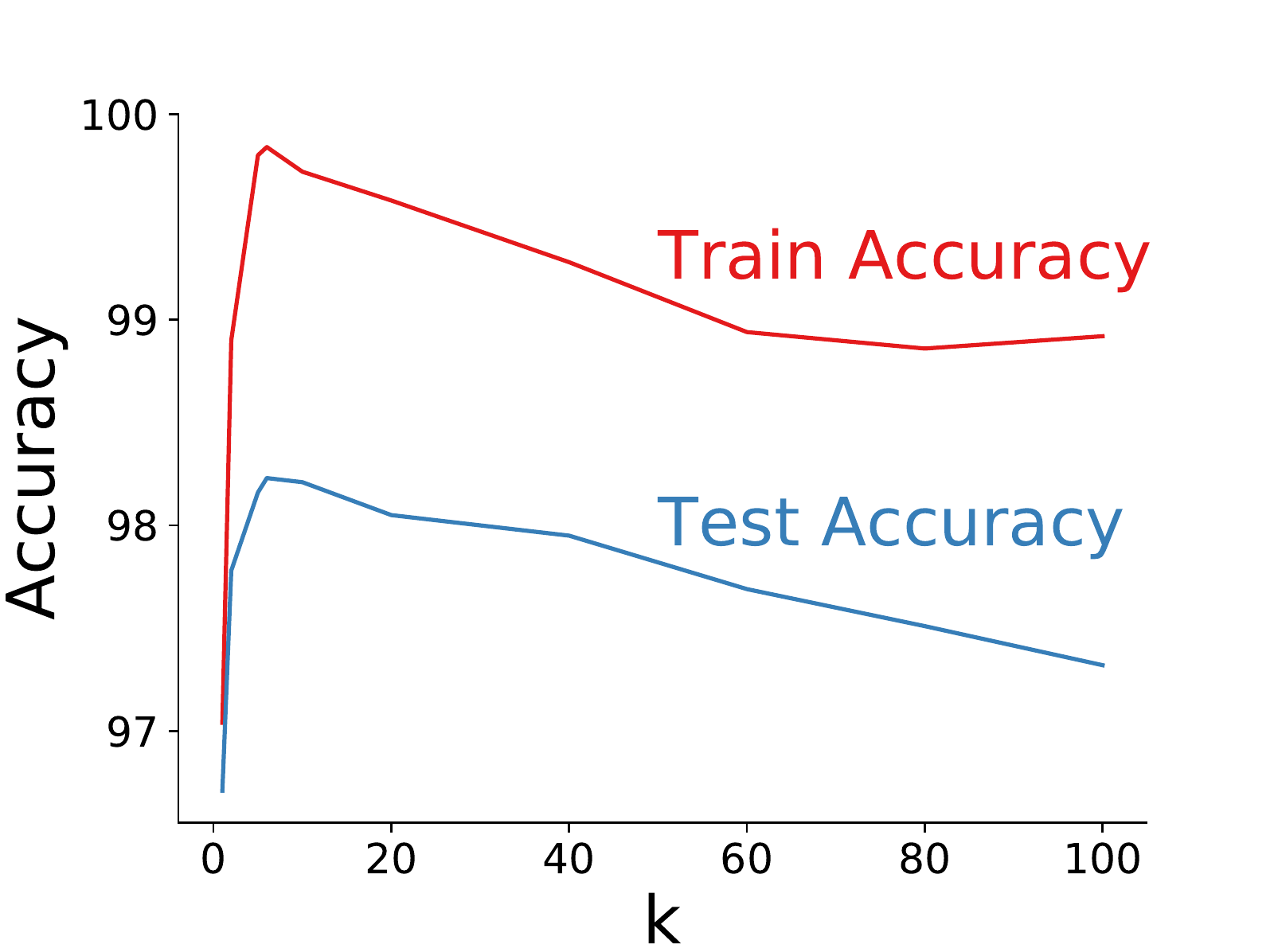}
    \label{fig:kc_opt}
    }
    \hfill
    \subfloat[]{
    \includegraphics[width=0.47\columnwidth]{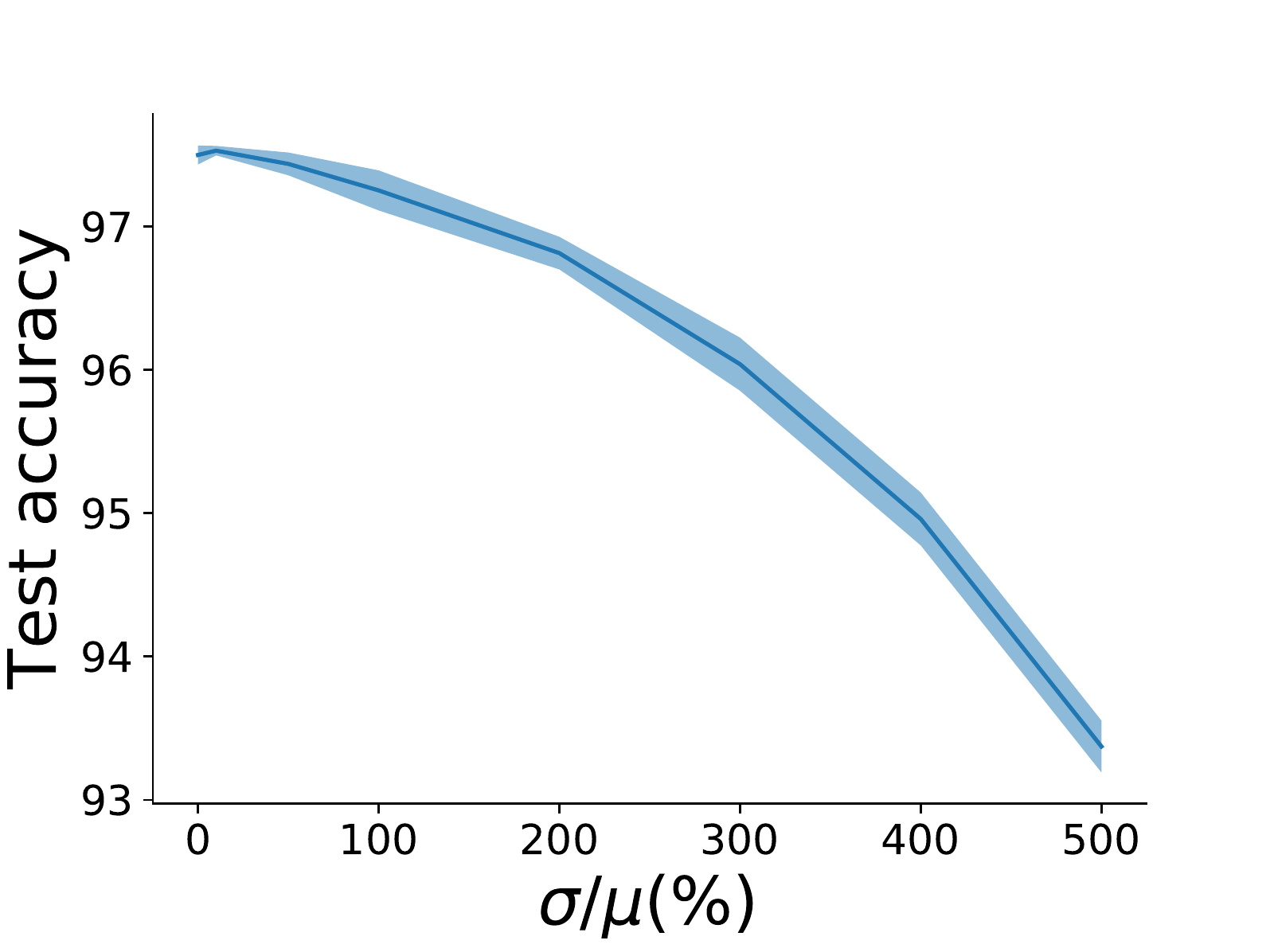}
    \label{fig:noise}
    }
    \caption{MNIST experiments: (a) Test accuracy as a function of the number of epochs on MNIST dataset (averaged over 10 runs, one standard error shaded). The difference in accuracy between floating point baseline and flash synapse RPU is negligible with 10\% noise and around 0.1\% with 100\% noise. (b) Train and test accuracy as a function of $k$ on the MNIST dataset. Accuracies are low for very low and very high values of $k$, with $k=6$ being the best value. (c) Test accuracy as a function of update noise on MNIST dataset after 3 epochs (averaged over 4 runs, one standard error shaded). The accuracy drops by less than 0.3\% with 100\% noise and by around 4\% with 500\% noise.}
\end{figure}

Figure \ref{fig:mnist} shows the learning curves with 10\% noise and 100\% noise respectively, compared with that of the baseline. The curves are averaged over the 10 runs and one standard error is shaded. 
The final accuracies with the flash device are $98.07 \pm 0.05\%$ and $97.91 \pm 0.06\%$ with 10\% noise and 100\% noise respectively. 
The final accuracy of the baseline is $98.05 \pm 0.07\%$.

\subsubsection{Effect of Weight Scaling Factor ($k$) on Performance}
\label{sec:k_performance}

As described in Section \ref{sec:flash_rpu}, changing $k$ leads to a trade-off between linearity, noise, and the number of pulses available. 
To study its effect on the performance, we adjust $k$ and measure the test and train accuracies. 

Figure \ref{fig:kc_opt} shows the variation of train and test accuracies for different values of k at a noise level of 10\%. 
The highest train accuracy of $99.84\%$ was obtained for $k=6$, with the corresponding test accuracy being $98.15\%$. 


Higher values of $k$ used a lower range of device conductances, which reduced the precision of the system since $\Delta_0^+(g)$ and $\Delta_0^-(g)$ are unchanged. 
Lower values of $k$ used a larger range of device conductances. 
Since the conductance change became more non-linear on either extreme, the performance declined.

\subsubsection{Noise Analysis}

In the above sub-sections, we showed plots for the flash device with a noise level of 10\% of the mean and 100\% of the mean. 
To further study the effect of noise on the performance, we run the MNIST experiments with noise level varying from 0\% to 500\% and find the test accuracy after 3 epochs. 

Figure \ref{fig:noise} shows the accuracy as a function of noise, averaged over 4 runs. 
The accuracy is $97.5 \pm 0.07\%$ without noise, $97.3 \pm 0.14\%$ with 100\% noise, and drops to $93.4 \pm 0.18\%$ at 500\% noise. 
As shown in section \ref{sec:exp_data}, 100\% noise is well above those found experimentally in the flash device, and hence, acts as a lower bound on the obtainable accuracy. 


\subsection{CIFAR}

CIFAR dataset consists of 50,000 training and 10,000 test images of real world objects. 
Each image is colored and 32x32 pixels in size. 
CIFAR-10 consists of 10 classes of images, while CIFAR-100 consists of 100 classes of images. 

Since convolutional neural networks (CNNs) are generally used for classification on these datasets, we follow the methodology used by  Ambrogio et al. \cite{ambrogio2018equivalent} to compare our device with the baseline. 
A pre-trained CNN, specifically, ResNet-50 \cite{resnet} pre-trained on the ImageNet \cite{imagenet} dataset, is used for feature extraction. 
The CIFAR images were resized, normalized and passed through the pre-trained network. 
The activations of the last hidden layer were considered as features.

Once the features were extracted, a neural network with no hidden layers was trained to classify the images based on the features. 
The neural network was trained for 10 epochs. 
Similar to the MNIST experiments, CIFAR experiments were repeated 10 times while recording test, train accuracies.

\begin{figure}[t]
    \centering
    \subfloat[CIFAR-10]{
    \includegraphics[width=0.47\columnwidth]{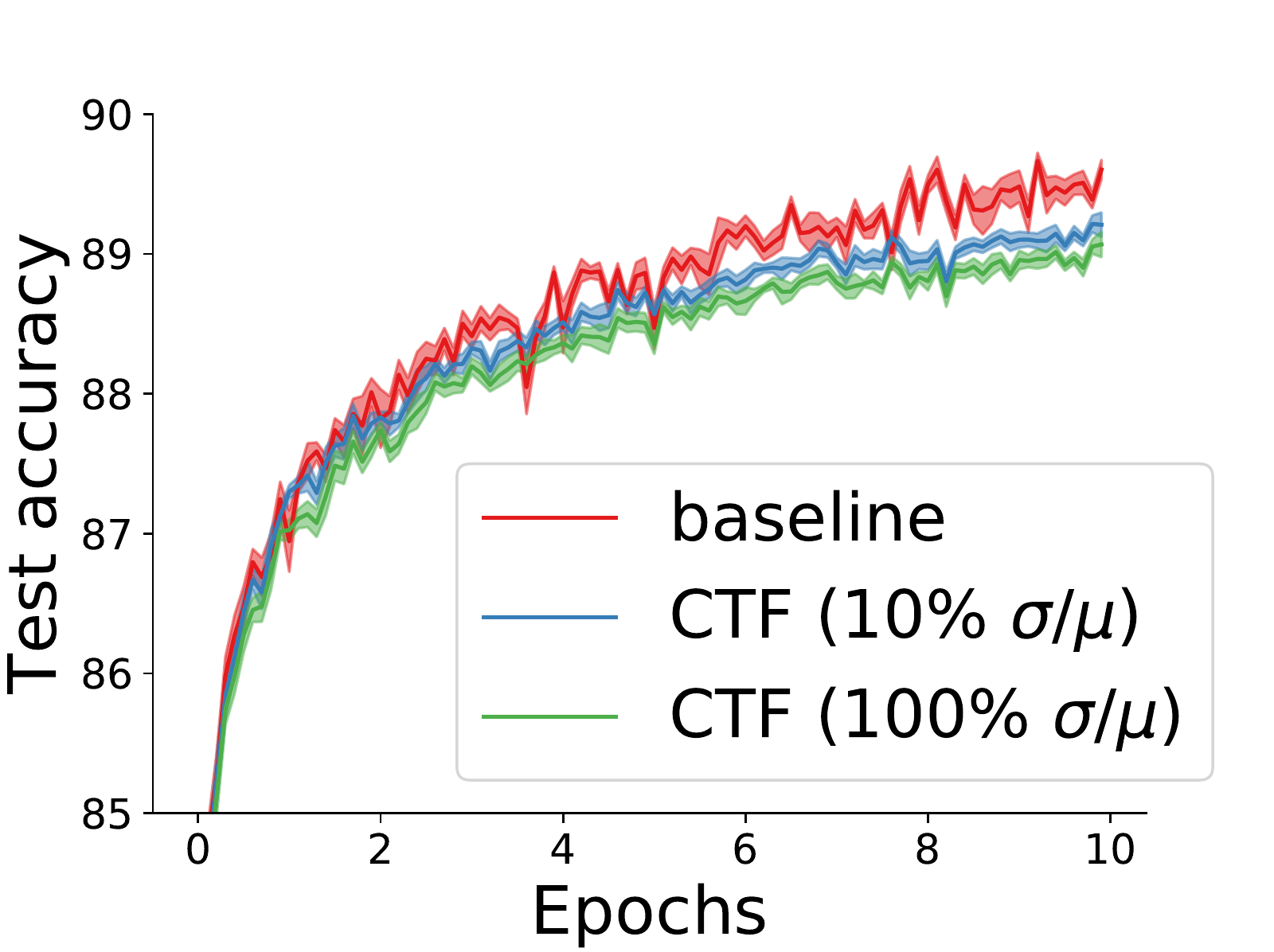}
    \label{fig:cifar10}
    }
    \hfill
    \subfloat[CIFAR-100]{
    \includegraphics[width=0.47\columnwidth]{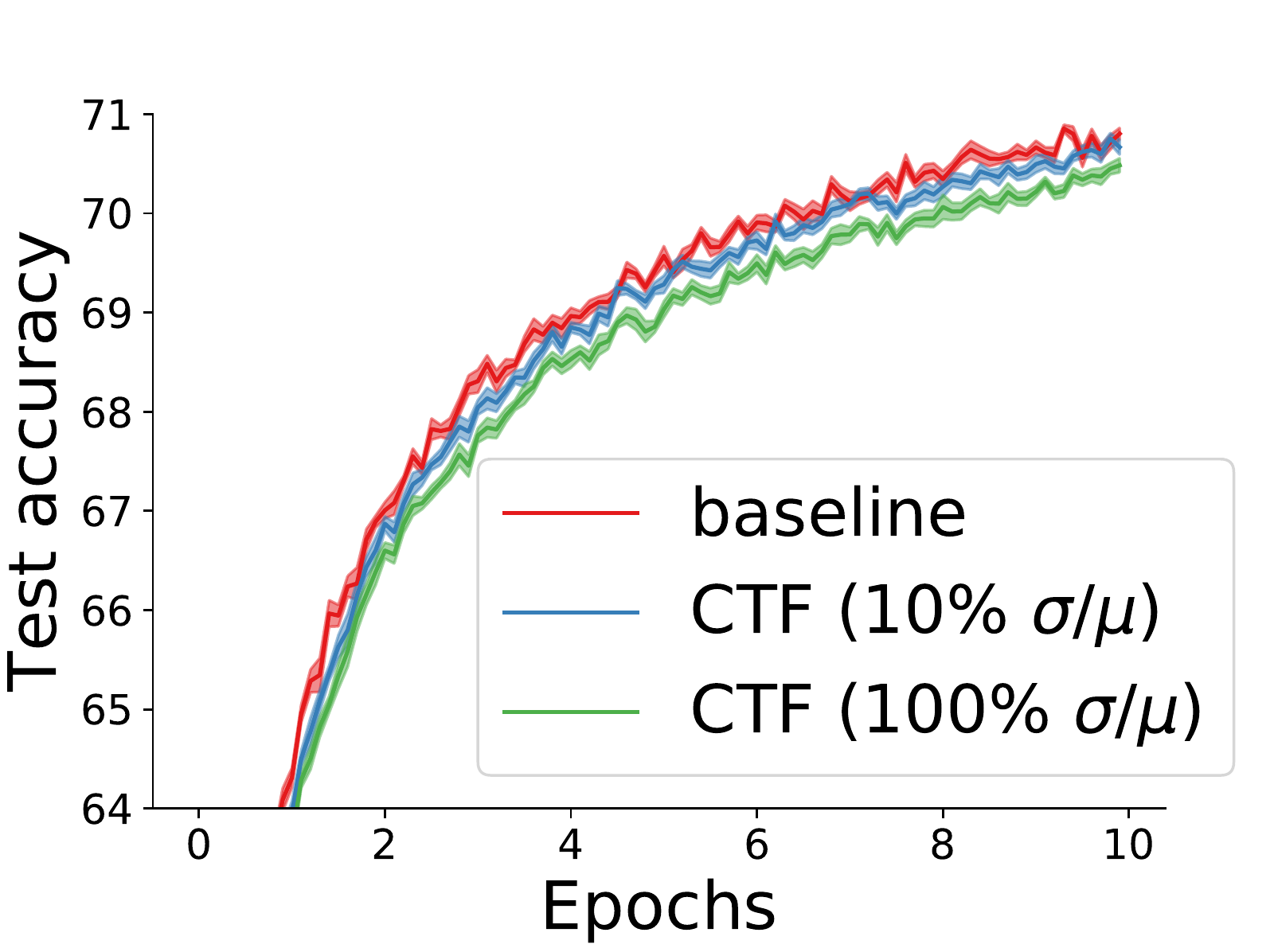}
    \label{fig:cifar100}
    }
    \hfill
    \subfloat[Mountain Car]{
    \includegraphics[width=0.47\columnwidth]{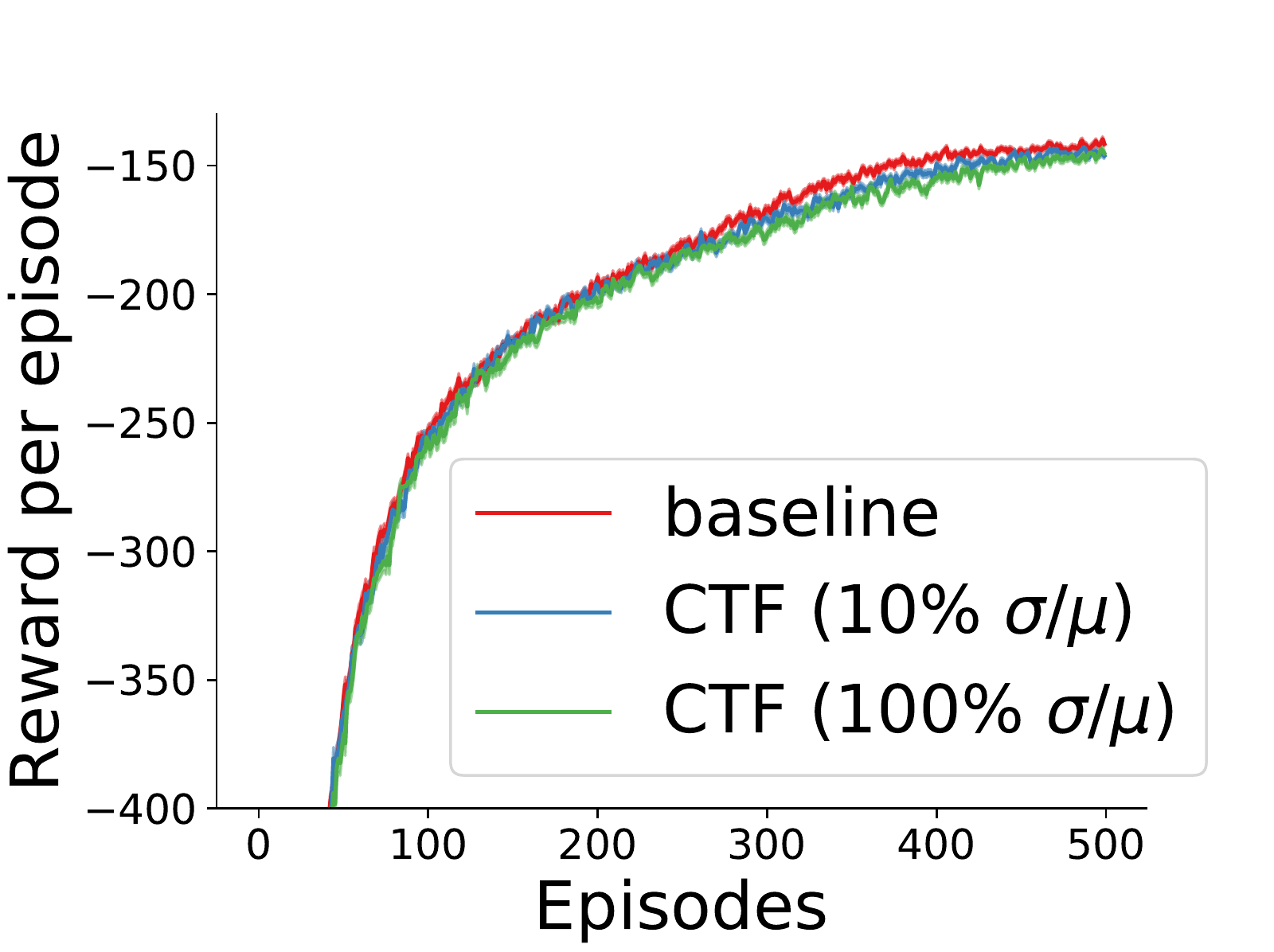}
    \label{fig:rl}
    }
    \caption{CIFAR and Mountain Car experiments: (a) Test accuracy as a function of the number of epochs on CIFAR-10 dataset (averaged over 10 runs, one standard error shaded). The difference in accuracy between floating point baseline and flash synapse RPU is around 0.4\% with 10\% noise and around 0.5\% with 100\% noise. (b) Test accuracy as a function of the number of epochs on CIFAR-100 dataset (averaged over 10 runs, one standard error shaded). The difference in accuracy between floating point baseline and flash synapse RPU is around 0.1\% with 10\% noise and around 0.3\% with 100\% noise. (c) Reward per episode as a function of the number of episodes on the Mountain Car environment (averaged over 100 runs, the standard error is less than the line width). The difference in reward between floating point baseline and flash synapse RPU is around 4\% with 10\% noise and around 3\% with 100\% noise.}
\end{figure}

Figure \ref{fig:cifar10} shows the learning curves with 10\% noise and 100\% noise respectively, for CIFAR-10 dataset. 
The final accuracies with the flash device are $89.21 \pm 0.09\%$ and $89.07 \pm 0.09\%$ respectively. 
The final accuracy of the baseline is $89.6 \pm 0.07\%$.

Figure \ref{fig:cifar100} shows the same for the CIFAR-100 dataset. 
The final accuracies with the flash device are $70.67 \pm 0.07\%$ and $70.48 \pm 0.07\%$ with 10\% and 100\% noise respectively.
The final accuracy of the baseline is $70.8 \pm 0.06\%$.


\begin{table*}[t]
    \caption{Comparison of our work with the previous works on MNIST dataset.}
    \label{tab:comparison}
    \centering
    \begin{tabularx}{\textwidth}{YYYC{1.5cm}C{1.4cm}Y}
        \toprule
        \textbf{Authors} & \textbf{Precision} & \textbf{Programming} & \textbf{Devices per Weight} & \textbf{MNIST Accuracy} & \textbf{Applications Demonstrated} \\
        \midrule
        Ambrogio et al. \cite{ambrogio2018equivalent} & Dual Precision: High precision, volatile DRAM + Low precision non-volatile PCM & Analog pulse V and time & 2 PCM + DRAM & \textbf{97.95\%} & Supervised Learning - MNIST, CIFAR-10, CIFAR-100 \\
        \midrule
        Nandakumar et al. \cite{nandakumar2018mixed} & Dual Precision: High precision, volatile CMOS + Low precision, non-volatile PCM & Analog pulse V and time & 2 PCM + SRAM & 97.40\% & Supervised Learning - MNIST \\
        \midrule
        Agarwal et al. \cite{agarwal2019using} & \textbf{Single precision} & Analog pulse V and time & \textbf{2 SONOS flash} & 97.6\% & Supervised Learning - File Types, MNIST \\
        \midrule
        Agarwal et al. \cite{agarwal2019using} & Dual Precision: High \& Low precision CTF by relative weight & Analog pulse V and time & 4 SONOS flash & \textbf{98\%} & Supervised Learning - File Types, MNIST \\
        \midrule
        Nandakumar et al. \cite{nandakumar2018phase} & \textbf{Single Precision} & \textbf{Stochastic Identical Pulse Train} & 2 PCM & 83\% & Supervised Learning - MNIST \\
        \midrule
        Babu et al. \cite{babu2018stochastic} & \textbf{Single Precision} & \textbf{Stochastic Identical Pulse Train} & \textbf{2 PCMO} & 88.1\% & Supervised Learning - MNIST \\
        \midrule
        This work & \textbf{Single Precision} & \textbf{Stochastic Identical Pulse Train} & \textbf{2 CTF} & \textbf{97.9\%} & \textbf{Supervised Learning - MNIST, CIFAR-10, CIFAR-100; Reinforcement Learning - Mountain Car} \\
        \bottomrule
    \end{tabularx}
\end{table*}

\subsection{Mountain Car}

Mountain Car is a control problem in which the agent should drive a car to the top of the mountain. 
The agent observes its current horizontal position (a real number between -1.2, 0.6) and its velocity (a real number between -0.07, 0.07). 
The goal is to reach the position of 0.5, which corresponds to the top of the peak. 
The agent can move forward, move backward or do nothing. 
Since the agent can’t accelerate enough to reach the peak by just moving forward, it needs to move back and forth to build enough momentum before being able to reach the peak \cite{mountain_car}. 
The agent gets a reward of -1 at every time step until it reaches the goal, and hence, it needs to reach the goal as quickly as possible.


We used tile coding \cite[pg.~217]{rl_book} to extract features from the observations and used a neural network with no hidden layers on top of it to predict the state-action values (Q-values) for each action. 
Mathematically, $\hat{q}(s,a; \bm{W})$ provided an approximation of $Q(s, a)$, for each state $s$ and action $a$. The weights were updated using Q-learning \cite{q_learning} update: 
\begin{align}
\begin{split}
    \bm{W} \leftarrow \bm{W} + \alpha(&R + \gamma \max_{a'}\hat{q}(S',a';\bm{W}) - \\
    &\hat{q}(S,A;\bm{W})) \nabla_{\bm{W}}\hat{q}(S,A;\bm{W})
\end{split}
\label{eq:q_learning}
\end{align}
where $S$ is the current state, $A$ is the action chosen, $R$ is the reward obtained, $S'$ is the next state, $\alpha$ is the step size, and $\gamma$ is the discount factor. The gradient calculation and weight update in Equation \ref{eq:q_learning} was performed by simulating the flash device.

Action selection was done using epsilon-greedy strategy with $\epsilon = 0.1$. 
Hash-based tile coding software by Sutton \cite{sutton_tile} was used for feature extraction, with 8 equally sized tiles per dimension and 16 tilings. 

The agent was trained for 500 episodes, with each episode being terminated either on reaching the goal or after 1000 steps. 
The experiment was repeated 100 times and the total reward obtained from each episode was recorded.

Figure \ref{fig:rl} shows the total reward per episode as a function of the number of episodes with 10\% noise and 100\% noise respectively. 
The floating point baseline obtains a reward of $-143 \pm 1.6$ (which implies that it takes around 143 steps to complete an episode). 
With the flash device, the reward is $-147 \pm 1.8$ with 10\% noise and $-146 \pm 2$ with 100\% noise.


\section{Discussions}
We show that the CTF device works as a replacement for floating point update in various applications.
In all the experiments, the performance of our device was close to that of the floating point baseline.
It was also fairly robust to the experimentally measured noise of 10-40\% in updates which is crucial for analog computing.

Classification on MNIST dataset showed that a multi-layer neural network can be trained using the CTF device. 
Classification on CIFAR-100 dataset showed that even in the regime of a large number of classes and relatively low data, the performance is on par with the floating point updates. 
Training an agent on Mountain Car environment showed that our method is not just restricted to the supervised learning setting, but can also be used in other settings that use neural networks. 


Table \ref{tab:comparison} shows that comparison of various current approaches.
Among various approaches for in-memory computing, precision enhancement of low precision but compact nanoscale memory like Phase Change Memory (PCM) with high precision but area inefficient CMOS memory enables high performance on MNIST dataset \cite{ambrogio2018equivalent,nandakumar2018mixed}. 
Further, single precision approaches with RPU based stochastic identical pulse based weight update show degraded performance of 83\% for PCM \cite{nandakumar2018phase} and of 88\% for PCMO based RRAM \cite{babu2018stochastic} on MNIST dataset.
Agarwal et al. \cite{agarwal2019using} have shown a single precision approach based on SONOS based Flash memory with analog pulse control with voltage and time to record a performance of 97.6\% on MNIST. 
This technology is based on NOR flash memory like programming scheme using high current/power technique of channel hot electrons (CHE).
Enhancing precision by a dual precision technique with more flash devices per weight and control circuit to enable a periodic carry improves MNIST performance to 98\%.

In comparison, our flash memory is programmed with the low current/power/energy FN tunneling technique.
Stochastic pulse train based RPU is demonstrated, eschewing the need for variable pulses with analog voltage levels and pulse time controls. 
The low rate of conductance change, high linearity produces a peak performance of 97.9\% - which is robust to experimentally measured noise levels. 
Further, our method produces excellent performance on various ANN applications like classification on CIFAR-10, CIFAR-100 datasets, and reinforcement learning on Mountain Car environment - demonstrating excellent generalization.

\section{Conclusions}



In this paper, we proposed a charge trap flash device in an RPU architecture to accelerate deep neural networks while maintaining software-level accuracy. 
The resistive processing unit speeds up vector-matrix and vector-vector multiplication operations, which are ubiquitously used in the backpropagation algorithm to train deep neural networks. 
We engineered the magnitude and the width of the pulse used to update the weights using the flash device. 
The updates were shown to be linear, gradual and symmetric, which is necessary for good performance.

We then simulated the device to train neural networks on MNIST, CIFAR-10 and CIFAR-100 datasets. 
In each case, the accuracy of the system was close to the floating point baseline, showing excellent generalization.
The system was also robust to noise in weight updates, with less than 1\% drop in accuracy when the simulated noise was 10x the experimentally observed value. 
We also demonstrated the generality of the method by applying it to a reinforcement learning method on the Mountain Car environment. 
The performance of our system matched the software baseline in this experiment too. 
Such implementation is benchmarked against the state of the art demonstrations to show best-in-class performance - indicating a promising hardware option for in-memory computing.

\bibliographystyle{IEEEtran}
\bibliography{IEEEabrv,references}

\end{document}